\documentclass[runningheads]{llncs}
\usepackage{graphicx}
\usepackage[english]{babel}
\usepackage[utf8]{inputenc}
\usepackage{csquotes}
\usepackage[noend]{algpseudocode}
\usepackage{amssymb}
\usepackage{booktabs}
\usepackage{multirow}
\usepackage{tabularx}
\usepackage{color}
\usepackage{booktabs}

\usepackage{amsmath,amsfonts,amssymb}
\usepackage{xspace} 

\usepackage{multirow}
\usepackage[normalem]{ulem}
\useunder{\uline}{\ul}{}

\usepackage[vlined,english,ruled,linesnumbered]{algorithm2e}

\newcommand{\ie}{i.e.}

\newcommand{\Real}{\mathop{\rm I\kern-.2emR}}

\renewcommand{\epsilon}{\varepsilon}

\newcommand{\set}[1]{\left\{{#1}\right\}}
\newcommand{\hide}[1]{}

\renewcommand{\set}[1]{\left\{{#1}\right\}}
\newcommand{\range}[2]{\set{{#1},\ldots,{#2}}}

\newcommand{\tcheb}{\textsf{g}}
\newcommand{\voisinage}{\mathcal B}

\newcommand{\EP}{\mathcal{EP}}

\newcommand{\DS}[0]{{X}}
\newcommand{\OS}[0]{{Z}}

\newcommand{\moead}{\textsc{Moea/d}\xspace}
\newcommand{\moeaddra}{\textsc{Moea/d--Dra}\xspace}
\newcommand{\moeadrnd}{\textsc{Moea/d--Rnd}\xspace}
\newcommand{\dra}{\textsc{Dra}\xspace}
\newcommand{\rnd}{\textsc{Rnd}\xspace}
\newcommand{\all}{\textsc{All}\xspace}

\newcommand{\spsall}{\textsf{sps}$_\all$\xspace}
\newcommand{\spsdra}{\textsf{sps}$_\dra$\xspace}
\newcommand{\spsrnd}{\textsf{sps}$_\rnd$\xspace}
\newcommand{\sps}{\textsf{sps}\xspace}

\newcommand{\hv}{\textsf{hv}\xspace}
\newcommand{\hvrd}{\textsf{hvrd}\xspace}

\begin{document}
\title{Study of MOEA/D population size in a multi-objective and rugged context}
\title{Analyzing the Effect of the Number of Objectives and of Ruggedness on the Population Size of MOEA/D}
\title{On the Effect of the MOEA/D Population Size for Problems with Different Objectives and Ruggedness}
\title{Analyzing the Effect of the MOEA/D Population Size on Ruggedness and Objective Space Dimension}
\title{On the Effect of the MOEA/D Population Size on Ruggedness and Objective Space Dimension}
\title{On the Combined Impact of Population Size and Sub-problem Selection in MOEA/D}

%
\titlerunning{Population Size and Sub-problem Selection in MOEA/D}
%
\author{Geoffrey~Pruvost\inst{1} 
\and
Bilel~Derbel\inst{1} 
\and
Arnaud~Liefooghe\inst{1} 
\and
Ke~Li\inst{2} 
\and
Qingfu~Zhang\inst{3}
}
%
\authorrunning{G. Pruvost et al.}

\institute{
University of Lille, CRIStAL, Inria, Lille, France,
\email{\{geoffrey.pruvost,bilel.derbel,arnaud.liefooghe\}@univ-lille.fr}\and
University of Exeter, Exeter, UK \\ \email{k.li@exeter.ac.uk} \and
City University Hong Kong, Kowloon Tong, Hong Kong \\ \email{qingfu.zhang@cityu.edu.hk}}
%
\maketitle              
\begin{abstract}
This paper intends to understand and to improve
the working principle of decomposition-based multi-objective evolutionary algorithms. 
We review the design of the well-established \moead framework to support the smooth integration of different strategies for sub-problem selection, while emphasizing the role of the population size and of the number of offspring created at each generation.
By conducting a comprehensive empirical analysis on a wide range of multi- and many-objective combinatorial NK landscapes, we provide new insights into the combined effect of those parameters on the anytime performance of the underlying search process. In particular, 
we show that even a simple random strategy selecting sub-problems at random outperforms 
existing sophisticated strategies. We also study the sensitivity of such strategies with respect to the ruggedness and the objective space dimension of the target problem.%
%
\end{abstract}
%
%
%
\section{Introduction}

\subsubsection{Context.} Evolutionary multi-objective optimization (EMO) algorithms~\cite{debMultiObjectiveOptimizationUsing2001} have been proved extremely effective in computing a high-quality approximation of the Pareto set, i.e., the set of solutions providing the best trade-offs among the objectives of a multi-objective combinatorial optimization problem (MCOP).
Since the working principle of
an evolutionary algorithm (EA)
is to evolve a \emph{population} of solutions, 
this population can be explicitly mapped with the target approximation set.
The goal is then to improve the quality of the population, 
and to guide its incumbent individuals to be as close and as diverse as possible w.r.t. the (unknown) Pareto set. 
Existing EMO algorithms
can be distinguished 
according to how the population is evolved. 
They are based on 
an
iterative process where at each iteration: (i)~some individuals (parents) from the population are selected, (ii)~new individuals (offspring) are generated using variation operators (e.g., mutation, crossover) applied to the selected parents, and (iii)~a replacement process 
updates
the population with newly generated individuals. Apart from the
problem-dependent
variation operators, 
the design of 
selection and replacement 
is well-understood to be the main challenge for an efficient and effective EMO algorithm, since these interdependent steps allow to control both the convergence of the population and its diversity. 
%
In contrast with dominance- (e.g.,~\cite{debMultiObjectiveOptimizationUsing2001})
or indicator- based (e.g.,~\cite{beume2007}) approaches,
aggregation-based approaches~\cite{moeadsurvey} 
rely on the transformation of the objective values of a solution into a 
scalar value, that can be used for selection and replacement. In this paper, we are interested in studying 
the working principles of this class of algorithms, while focusing on the so-called \moead (Multi-objective evolutionary algorithm based on decomposition)~\cite{LZ09,ZL07}, which can be considered as a state-of-the-art framework. 

\subsubsection{Motivations.} The \moead framework is based on the decomposition of the original MCOP into a set of 
smaller sub-problems that are mapped to a population of individuals.
In its basic variant~\cite{ZL07}, \moead considers a set of single-objective sub-problems defined using a scalarizing function transforming a multi-dimensional objective vector into a scalar value w.r.t. one weight (or direction) vector in the objective space. The population is then typically structured by mapping one individual to one 
sub-problem targeting a different region of the objective space. Individuals from the population are 
evolved following a cooperative mechanism 
in order for each individual (i)~to optimize its own 
sub-problem, and also (ii)~to help solving its neighboring sub-problems.
The population hence ends up having a good quality w.r.t. all sub-problems.
Although being extremely simple and flexible, the computational flow
of
\moead is constantly redesigned 
to deal with different issues. 
Different \moead variants have been proposed so far in the literature, e.g., 
to study the impact of elitist replacements~\cite{adaptive}, of generational design~\cite{ppsn2014}, or of stable-matching based evolution~\cite{keliStableMatchingBasedSelection2014}, and other mechanisms~\cite{Aghabeig2019}. In this paper, we are 
interested in the interdependence between the population size, which is implied by the number of 
sub-problems defined in the initial decomposition, and the internal evolution mechanisms of \moead. 

The population size has a deep impact on the dynamics and performance of EAs. 
In
\moead, the sub-problems target diversified and representative regions of the Pareto front. They are usually defined to spread evenly in the objective space. Depending on the shape of the (unknown) Pareto front, and 
on
the number of objectives, one may need to define a different number of sub-problems. Since, the population is structured following the so-defined sub-problems, it is not clear how the robustness of the \moead selection and replacement strategies
can be impacted by a particular setting of the population size.
Conversely, it is not clear what population size shall be chosen, and how to design a selection and replacement mechanism implying a high-quality approximation. Besides, the proper setting of the population size (see e.g.~\cite{crepinsekExplorationExploitationEvolutionary2013,glasmachersStartSmallGrow2014,wittPopulationSizeRuntime2008}) in is EAs can depend on the problem properties,
for example in terms of solving difficulty.
EMO algorithms are no exceptions.
In
\moead, 
sub-problems may have different characteristics, and the selection and replacement mechanisms can be guided by such considerations. This is for example the case for a number of \moead variants where it is argued that some sub-problems might be more difficult to solve than others~\cite{wangNewResourceAllocation2019,subprobEquallyImportant2016}, and hence that 
the population shall be guided accordingly.

\subsubsection{Methodology and contribution.} In this paper, we rely on the observation that the guiding principle of \moead can be leveraged in order to support a simple and high-level tunable design of the selection and replacement mechanisms on one hand, while enabling a more fine-grained control over the choice of the population size, and subsequently its impact on approximation quality on the other hand. More specifically, our work can be summarized as follows:%
%
\begin{itemize}
\item We consider a revised design of \moead which explicitly dissociates between three components: (i) the number of individuals selected at each generation, (ii) the strategy adopted for selecting those individuals and (iii) the setting of the population size. Although some sophisticated strategies to distribute the computational effort of sub-problems exploration were integrated within some \moead variants~\cite{claus,wangNewResourceAllocation2019,subprobEquallyImportant2016}, to the best of our knowledge, the individual impact of such components were loosely studied in the past.
\item Based on this fine-grained revised design, we conduct a comprehensive 
analysis about the impact of those three components on the convergence profile of \moead. Our analysis is conducted in an 
incremental manner, with the aim of providing insights about the interdependence between those design components. In particular, we show evidence that the number of sub-problems selected at each generation 
plays an even more important role than the way the sub-problems are selected. Sophisticated selection strategies from the literature are shown to be outperformed by simpler, well configured strategies.%
\item We consider a broad range of multi- and many-objective NK landscapes, viewed as a standard and difficult family of MCOP benchmarks, which is both scalable in the number of objectives and exposes a controllable difficulty in terms of ruggedness. By a 
thorough
benchmarking effort, we are then able to better elicit the impact of the \moead population size, and 
the robustness of selection strategies on the (anytime) approximation quality.
\end{itemize}
%
It is worth noticing that our work shall not be considered as yet another variant in the \moead literature. In fact, our analysis precisely aims at enlightening the main critical design parameters and components that can be hidden behind a successful \moead setting. Our investigations are hence to be considered as a step towards the establishment of a more advanced component-wise configuration methodology allowing the setting up of future high-quality decomposition-based EMO algorithms for both multi- and many-objective optimization.%

\subsubsection{Outline.} In Sec.~\ref{sec:background}, we recall basic 
definitions and we detail the working principle of \moead. In Sec.~\ref{sec:framework}, we describe our 
contribution in rethinking \moead by explicitly dissociating between the population size and the number of selected sub-problems, 
then allowing us to leverage existing algorithms
as 
instances of the revised 
framework. In Sec.~\ref{sec:results}, we present 
our experimental study and we state our main 
findings. In Sec.~\ref{sec:conclusion}, we conclude the paper and discuss 
further research.

\section{Background}
\label{sec:background}

\subsection{Multi-objective Combinatorial Optimization}
A \emph{multi-objective combinatorial optimization problem} (MCOP) can be defined by a set of $M$ objective functions $f=(f_1, f_2,\dots, f_M)$,
and a discrete set $\DS$ of feasible solutions in the \emph{decision space}.
Let $\OS  =  f(\DS) \subseteq \Real^M$ be the set of feasible outcome vectors in the  \emph{objective space}. To each solution $x \in \DS$ is assigned an objective vector $z \in \OS$, on the basis of the vector function \mbox{$f : \DS \rightarrow \OS$}.
In a maximization context, an objective vector $z \in \OS$ is \emph{dominated} by a vector $z^\prime \in \OS$ 
iff $\forall m \in \{1,\dots,M\}$, $z_m \leqslant z_m^\prime$ and $\exists m \in \{1,\dots,M\} $ s.t. $z_m < z_m^\prime$. A solution $x \in \DS$ is dominated by a solution $x^\prime \in \DS$ 
iff $f(x)$ 
is dominated by $f(x^\prime)$. A solution $x^\star \in \DS$ is \emph{Pareto optimal} if there does not exist any other solution $x \in \DS$ such that $x^\star$ 
is dominated by $x$.
The set of all Pareto optimal solutions is the \emph{Pareto set}. Its mapping in the objective space is the \emph{Pareto front}. 
The size of the Pareto set is typically exponential in the problem size.
Our goal is to identify a good \emph{Pareto set approximation}, for which 
EMO algorithms constitute a popular effective option~\cite{debMultiObjectiveOptimizationUsing2001}. As mentioned before, we are interested in aggregation-based methods, and especially in the \moead framework which is sketched below.

\subsection{The Conventional MOEA/D Framework}
Aggregation-based EMO algorithms seek good-per\-forming solutions in multiple regions of the Pareto front by \emph{decomposing} the original multi-objective problem into a number of \emph{scalarized} single-objective \emph{sub-problems}~\cite{moeadsurvey}. 
In this paper, we use the 
Chebyshev 
scalarizing function: 
$
\tcheb(x, \omega)	= \max_{i \in \set{1, \ldots, M}} \omega_i \cdot \big| z^\star_i - f_i(x) \big|
$,
where $x\in X$, $\omega = (\omega_1, \ldots, \omega_M)$ is a positive weight vector,
and $z^\star = (z^\star_1, \ldots, z^\star_M)$ is a reference point such that $z^\star_i > f_i(x)$
$\forall x \in \DS$, $i \in \set{1, \ldots, M}$.

In \moead~\cite{ZL07}, sub-problems are optimized cooperatively by defining a \emph{neighborhood relation} between sub-problems. Given a set of $\mu$ weight vectors $\mathcal{W}_\mu=(\omega^1, \ldots, \omega
^\mu)$, with $\omega^j=(\omega^j_1, \ldots, \omega^j_M)$ for every $j\in \range{1}{\mu}$, defining $\mu$ sub-problems, \moead maintains a population $P_\mu=(x^1, \ldots, x^\mu)$ where each individual $x^j$ corresponds to one sub-problem. For each sub-problem $j \in \set{1, \ldots, \mu}$, a set of neighbors $\voisinage_j$ is defined by considering the $T$ closest weight vectors based on euclidean distance. 
%
%
Given a sub-problem~$j$, two sub-problems are selected at random from $\voisinage_j$, and the two corresponding solutions are considered as parents. 
An offspring $x^\prime$ is created by means of variation (e.g., crossover, mutation). For every $k \in \voisinage_j$, if $x^\prime$ improves $k$'s current solution $x^k$, then $x^\prime$ replaces it, \ie, if $\tcheb(x^\prime,\omega^k) < \tcheb(x^j,\omega^k)$ then $x^k = x^\prime$. The algorithm loops over sub-problems, i.e., weight vectors, or equivalently over the individuals in the population, until a stopping condition is satisfied. In the conventional \moead terminology, an iteration refers to making selection, offspring generation, and replacement 
for
\emph{one} sub-problem. By contrast, a generation consists in processing all sub-problems once, i.e., after one generation $\mu$ offspring are generated. 
Notice that other
issues are 
also addressed,
such as the update of the reference point~$z^\star$ required by the scalarizing function, and the option to incorporate an external archive for storing all non-dominated points found so far during the search process.

From the previous description, it should be clear that, at each iteration, \moead is applying an elitist $(T+1)$-EA w.r.t. the sub-population $\voisinage_i$ underlying the neighborhood of the current sub-problem. After one generation, one can roughly view \moead as applying a 
$(\mu+\mu)$-EA w.r.t. the full population. A noticeable difference is that the basic \moead is not a generational algorithm, in the sense that it does not handle the population as a whole, but rather in a local and greedy manner. 
This is actually 
a
distinguishable feature of \moead, since the population is structured 
by
the initial sub-problems and evolved accordingly.%

\section{Revising and Leveraging the Design of MOEA/D}
\label{sec:framework}

\subsection{Positioning and Rationale}
As in any EA, both the population size and the selection and replacement mechanisms of \moead play a crucially important role. Firstly, a number of weight, i.e., a population size, that is too small may not only be insufficient to cover well the whole Pareto front, but may also prevent the identification of high-quality solutions for the defined sub-problems. This is because the generation of new offspring is guided by the so-implied $(T+1)$-EA for which the local sub-population of size $T$ might be too different and hence too restrictive for generating good offspring. 
On the other hand, a too large population may result in a substantial waste of resources, since too many sub-problems might map to the same solution.
Secondly, a small population size can be sufficient to approach the Pareto front in a reduced number of steps. However, a larger population 
is preferable to better cover the Pareto front.
%
As in single-objective optimization, a larger population might also help escaping local optima~\cite{wittPopulationSizeRuntime2008}. As a result, it is not clear what is a proper setting of the population size in \moead, since the previously discussed issues seem contradictory. 

Although one can find different studies dealing with the impact of the population size in EAs~\cite{crepinsekExplorationExploitationEvolutionary2013,glasmachersStartSmallGrow2014,wittPopulationSizeRuntime2008,corusStandardSteadyState2018},
this issue is explicitly studied only to a small extent, especially for decomposition-based multi- and many-objective optimization~\cite{TanabeI18,twolayer}. For instance, in~\cite{glasmachersStartSmallGrow2014}, offline and online scheduling strategies for controlling the population size are coupled with SMS-EMOA~\cite{beume2007}, a well-known indicator-based EMO algorithm, for bi-objective continuous benchmarks. Leveraging such a study to combinatorial domains with more than two objectives, and within the \moead framework, is however a difficult question. Tightly related to the population size, other studies investigate the distribution of the computational effort over the sub-problems~\cite{CaiLFZ15,chiangMOEADAMSImproving2011,claus,wangNewResourceAllocation2019,subprobEquallyImportant2016}. 
The rationale is that the defined sub-problems might have different degrees of difficulty and/or that the progress over some sub-problems might be more advanced than others in the course of the search process. Hence, different adaptive mechanisms have been designed in order to detect which sub-problems to consider, or equivalently which solutions to select when generating a new offspring. 
A representative example of such approaches is the so-called \moeaddra (\moead with dynamical resource allocation)~\cite{subprobEquallyImportant2016}, 
that can be considered as a state-of-the-art algorithm when dealing with the proper distribution of the computational effort over sub-problems. In \moeaddra, a utility function is defined w.r.t. the current status of sub-problems. A tournament selection is used to decide which sub-problems to select when generating a new offspring. Despite a skillful design, such an approach stays focused on the relative difficulty of solving sub-problems, while omitting to analyze the impact of the number of selected sub-problems and its interaction with both the population size and the characteristics of the underlying~problem.%

We propose to revise the \moead framework and to study in a more explicit and systematic manner the combined effect of population size and sub-problem selection in light of the properties of the MCOP at hand. As mentioned above,
this was investigated only to a small extent in the past although, as revealed by our experimental findings, it is of critical importance to reach an optimal performance when adopting the \moead framework.

\subsection{The Proposed MOEA/D--($\mu$, $\lambda$, \textsf{sps}) Framework}
In order to better study and analyze the combined effect of population size and sub-problem selection, we propose to rely on a revised framework for \moead, denoted \moead--($\mu$, $\lambda$, \textsf{sps}), as defined in the high-level template depicted in Algorithm~\ref{algo:moeadsps}. This notation is inspired by the standard $(\mu+\lambda)$-EA scheme, where starting from a population of size $\mu$, $\lambda$ new individuals are generated and merged to form a new population of size $\mu$ after replacement. In the \moead framework, however, this has a specific meaning as detailed in the following.

The proposed \moead--($\mu$, $\lambda$, \textsf{sps}) algorithm follows the same steps as the original \moead. However, it explicitly incorporates an additional component, denoted \textsf{sps}, which stands for the \underline{\textbf{s}}ub-\underline{\textbf{p}}roblem \underline{\textbf{s}}election strategy. Initially, the population is generated and mapped to the initial $\mu$ weight vectors.
An optional external archive 
is also incorporated in the usual way with no effect on the search process.
%
The algorithm then proceeds in different generations.
At each generation, $\lambda$ sub-problems, denoted $I_\lambda$, are selected using the \textsf{sps} strategy. A broad range of deterministic and stochastic selection strategies can be integrated. In particular, $\lambda$ can be though as an intrinsic parameter of the EMO algorithm itself, or implied by a specific \textsf{sps} strategy. The so-selected sub-problems are processed in order to update the population. 
For the purpose of this paper, we adopt the same scheme than conventional \moead:
selected sub-problems are 
processed in an iterative manner,
although other generational EA schemes could be adopted.
%
At each iteration, that is for each selected sub-problem, denoted $i$, some parents are selected as usual from the $T$-neighborhood $\voisinage_i$ of weight vector $\omega^i$ w.r.t. $\mathcal{W}_\mu$. The setting of the neighborhood $\voisinage_i$ can be exactly the same as in conventional \moead and its variants. However, at this step, it is important to emphasize that 
$\voisinage_i$ may include some sub-problems that were \emph{not} selected by the \textsf{sps} strategy. This is motivated by the fact that parents that are likely to produce a good offspring should be defined w.r.t. the population as a whole, 
and not solely within the subset of active sub-problems at a given generation, which might be restrictive.
%
A new offspring $x^\prime$ is then generated using standard variation operators (e.g., crossover, mutation).
The reference point required by the scalarizing function and the optional external archive are updated. 
Thereafter, the offspring is considered for replacement as in the conventional \moead and its variants. Here again, this is handled using the neighborhood~$\voisinage_i$ of the current sub-problem $i$, computed w.r.t. the whole population. It is worth noticing that population update is made on the basis of the scalarizing function~$\tcheb$, which is a 
distinguishable feature of aggregation-based approaches.

At last, notice that we also use a \emph{history} variable, 
serving as a memory where
any relevant information could be store
for the future 
actions of the algorithm. 
In particular, we explicitly integrate the history within the \textsf{sps} strategy, since this will allow us to leverage some existing \moead variants.

\begin{algorithm}[!t]
	\caption{High level template of \moead--($\mu$, $\lambda$, \textsf{sps})}\label{algo:moeadsps}
\SetKwInput{KwInput}{Input}
 \KwInput{$\mathcal{W}_{\mu} := \set{\omega^1, \ldots, \omega^\mu}$: weights; $\tcheb(\cdot ~|~\omega)$: scalar function; $T$: neighb. size;}

$\EP\gets \varnothing$ : (optional) external archive \;
$\mathcal{P}_\mu \gets \set{x^1, \ldots, x^\mu}$ : generate and evaluate initial population of size $\mu$\;
$z^\star\gets$ initialize reference point from $\mathcal{P}_\mu$\;
		
\While{$Stopping Criteria$}{
	$I_\lambda \gets \textsf{sps}(\mathcal{W}_{\mu}, \mathcal{P}_{\mu}, history)$\;
	\For{$i \in I_\lambda$}{
		$\voisinage_i \gets$ the $T$-neighborhood of sub-problem $i$ using $\mathcal{W}_\mu$\;
		$\mathcal{X}\gets \textrm{matingSelection}(\voisinage_i)$\;
		$x^\prime \gets \textrm{variation}(\mathcal{X})$\;
		$F(x^\prime) \gets$ evaluate $x^\prime$ \; 
		$\EP \gets$ update external archive using $x'$\;
		$z^\star\gets$ update reference point using $F(x^\prime)$\;
		$\mathcal{P}_\mu \gets \textrm{replacement}(\mathcal{P}_\mu,x^\prime,\voisinage_i ~|~\tcheb)$\;
		
		$history \gets$ update search history\;
	}
}
\end{algorithm}

\subsection{Discussion and Outlook}
It shall be clear from the previous description that the \moead--($\mu$, $\lambda$, \textsf{sps}) framework allows us to emphasize the interdependence between three main components in a more fine-grained manner while following the same working principle than the original \moead. 
Firstly, the number of weight vectors, or equivalently the population size, is now made more explicit. In fact, the set of weight vectors now `simply' plays the role of a global data structure to organize the individuals from the population. This structure can be used at the selection and replacement steps. 
In particular, one is not bound to iterate over all weight vectors, but might instead select a subset of individuals following a particular strategy.
Secondly, the number of selected sub-problems~$\lambda$ determines directly the number of offspring to be generated at each generation. From an exploration/exploitation perspective, we believe this is of critical importance in general for $(\mu+\lambda)$-EAs, and it is now made more explicit within the \moead framework. Furthermore, the $\lambda$ offspring solutions are not simply generated from the individuals mapping to the selected sub-problems. Instead, parent selection interacts directly with the whole population, structured around the $\mu$ weight vectors, since the local neighborhood of each selected sub-problem may be used. 
Thirdly, the interaction between $\mu$ and~$\lambda$ is complemented more explicitly by the sub-problem selection strategy. In conventional \moead for instance, the selection strategy turns out to be: \textsf{sps}$_\all$ = `select all sub-problems', with $\lambda=\mu$. 
However, advanced \moead variants can be captured as well. For instance, \moeaddra~\cite{subprobEquallyImportant2016}, focusing on the dynamic distribution of computations, can easily be instantiated 
as follows. For each sub-problem, we store and update the utility value as introduced in~\cite{subprobEquallyImportant2016} by using the history variable. Let us recall that in \moeaddra, the utility of a sub-problem is simply the amount of progress made by solution $x^i$ for sub-problem~$\omega^i$ in terms of the scalarized fitness value $\tcheb(\cdot | \omega^i)$ over different generations. In addition, $M$ boundary weight vectors (in the objective space) are selected 
at each generation, and further $(\mu/5 - M)$ weight vectors are selected by means of a tournament selection of size $10$. Hence, the sub-problem selection strategy turns out to be \textsf{sps}$_\dra$ = `select the boundary vectors and sub-problems using a tournament selection of size $10$', with $\lambda=\mu/5$. Notice that 
this choice
is to recall the 
one-fifth success rule from $(\mu+\lambda)$ evolution strategies~\cite{onefifth}.

In the reminder, \moead--($\mu$, $\mu$, \textsf{sps}$_\all$) refers to the conventional \moead as described in~\cite{ZL07}, and  \moead--($\mu$, $\mu/5$, \textsf{sps}$_\dra$) refers to \moeaddra~\cite{subprobEquallyImportant2016}; see
Table~\ref{tab:algos}. Other 
parameters can be conveniently investigated as well. Since we are interested in the combined effect of $\mu$, $\lambda$ and \textsf{sps}, 
we also consider a simple baseline sub-problem selection strategy, denoted \textsf{sps}$_\rnd$, which is to select a subset of sub-problems uniformly at random. Notice that 
our empirical analysis shall shed more lights on the behavior and the  accuracy of the existing \textsf{sps}$_\dra$ strategy. 

\begin{table}[!t]
\centering
\caption{Different instantiations of the \moead-$(\mu,\lambda,\textsf{sps})$ framework.
}
\label{tab:algos}
\begin{tabular}{l|c|c|c|l}
\toprule
\textbf{~algorithm} & \textbf{~pop. size~} & \textbf{~\# selected sub-prob.~} & \textbf{~selection strategy~} & \textbf{~ref. ~}\\
\midrule
~\moead & $\mu$ & $\mu$ &  \textsf{sps}$_\all$ & ~\cite{ZL07} \\
~\moeaddra & $\mu$ & $\mu/5$ & \textsf{sps}$_\dra$ & ~\cite{subprobEquallyImportant2016} \\
~\moeadrnd~ & $\mu$ & $\lambda \leqslant \mu$ & \textsf{sps}$_\rnd$ & ~here~ \\
\bottomrule
\end{tabular}
\vspace{-2ex}%
\end{table}

\section{Experimental Analysis}
\label{sec:results}

\subsection{Experimental Setup}

\subsubsection{Multi-objective NK landscapes.}
We consider multi-objective NK landscapes as a  problem-independent model of multi-objective multi-modal combinatorial optimization problems~\cite{verelStructureMultiobjectiveCombinatorial2013}. 
Solutions are binary strings of size~$N$ and
the objective 
vector to be maximized 
is defined as $f\colon \lbrace 0, 1 \rbrace^{N} \mapsto [0,1]^M$. 
%
The parameter $K$ defines the \emph{ruggedness} of the problem,
that is the number of (random) variables that influence the contribution of a given variable to the objectives.
%
By increasing 
$K$ from $0$ to~$(N-1)$, problems can be gradually tuned from smooth to rugged.
%
We consider instances with the following settings:
the problem size is set to $N=100$, the number of objectives to $M \in \{2,3,4,5\}$, and the ruggedness to $K \in \{0, 1, 2, 4\}$, that is, from linear to highly rugged landscapes. We generate one instance 
at random for each combination. 

\subsubsection{Parameter setting.}
For our analysis, we consider three competing algorithms extracted from the \moead--($\mu$, $\lambda$, $\textrm{sps}$) framework as depicted in Table~\ref{tab:algos}. For the conventional \moead, only one parameter is kept free, that is the population size $\mu$. For \moeaddra, the sub-problem selection strategy is implemented as described in the original paper~\cite{subprobEquallyImportant2016}. We further consider to experiment \moeaddra with other $\lambda$ values. Recall that in the original variant, only $\mu/5$ sub-problems are selected, while including systematically the $M$ boundary weight vectors. For fairness, we follow the same principle when implementing the \moeadrnd strategy. Notice that the boundary weight vectors were shown to impact the coordinates of the reference point $z^\star$ used by the scalarizing function~\cite{wangNewResourceAllocation2019}. They are then important to consider at each generation. To summarize, for both \moeaddra and \moeadrnd, two parameters are kept free, namely the population size $\mu$ and the number of selected sub-problems $\lambda$. They are chosen to cover a broad range of values, from very small to relatively very high,
namely, $\mu \in \set{1, 10, 50, 100, 500}$ and $\lambda \in \set{1, 2, 5, 10, 25, 50, 100, 150, 200, 300, 400, 450, 500}$ 
such that $\lambda \leqslant \mu$. 

The other common parameters are set as follows. The initial weights are generated using the 
methodology described in~\cite{zapotecas-martinezLowdiscrepancySequencesTheir2015}. The neighborhood size is set to $20\%$ of the population size: $T = 0.2 \, \mu$. Two parents are considered for mating selection, i.e., the parent selection in the neighborhood of a current sub-problem~$i$. The first parent is the current solution $x^i$, and the second one is selected uniformly at random from $\voisinage_i$. Given that solutions are binary strings, we use a two-point crossover operator and a bit-flip mutation operator where each bit is flipped with a rate of $1/N$. 
\moeaddra involves additional parameters which are set 
following the recommendations from~\cite{subprobEquallyImportant2016}.%

\subsubsection{Performance evaluation.}
Given the large number of parameter values
(more than $2\,000$ different configurations), and in order to keep our experiments manageable in a reasonable amount of time, every configuration is executed $10$ independent times, for a total of 
more than $20\,000$
runs. In order to appreciate the convergence profile and the anytime behavior of the competing algorithms, we consider different stopping conditions of $\set{10^0, 10^1, \ldots, 10^7}$ calls to the evaluation function.  Notice however that due to lack of space, we shall only report our findings on the basis of a representative set of our experimental data.
%

For performance assessment, we use the hypervolume indicator (\hv)~\cite{zitzler2003} to assess the quality of the obtained approximation sets, the reference point being set to the origin. More particularly, we consider the hypervolume relative deviation, computed as $\hvrd(A) = (\hv(R) - \hv(A)) / \hv(R)$, where $A$ is the obtained approximation set, and $R$ is the best Pareto front approximation, obtained by aggregating the results over all executions and removing dominated points. 
As such, a lower value is better.
It is important to notice that we consider the external archive, storing all non-dominated points found so far during the search process, for performance assessment. This is particularly important when comparing configurations using different population sizes. 

\subsection{Impact of the Population Size: \textsf{sps}$_\all$ with Varying $\mu$ Values}
\label{sec:popmoead}

\begin{figure}[!t]
	\includegraphics[width=1\linewidth]{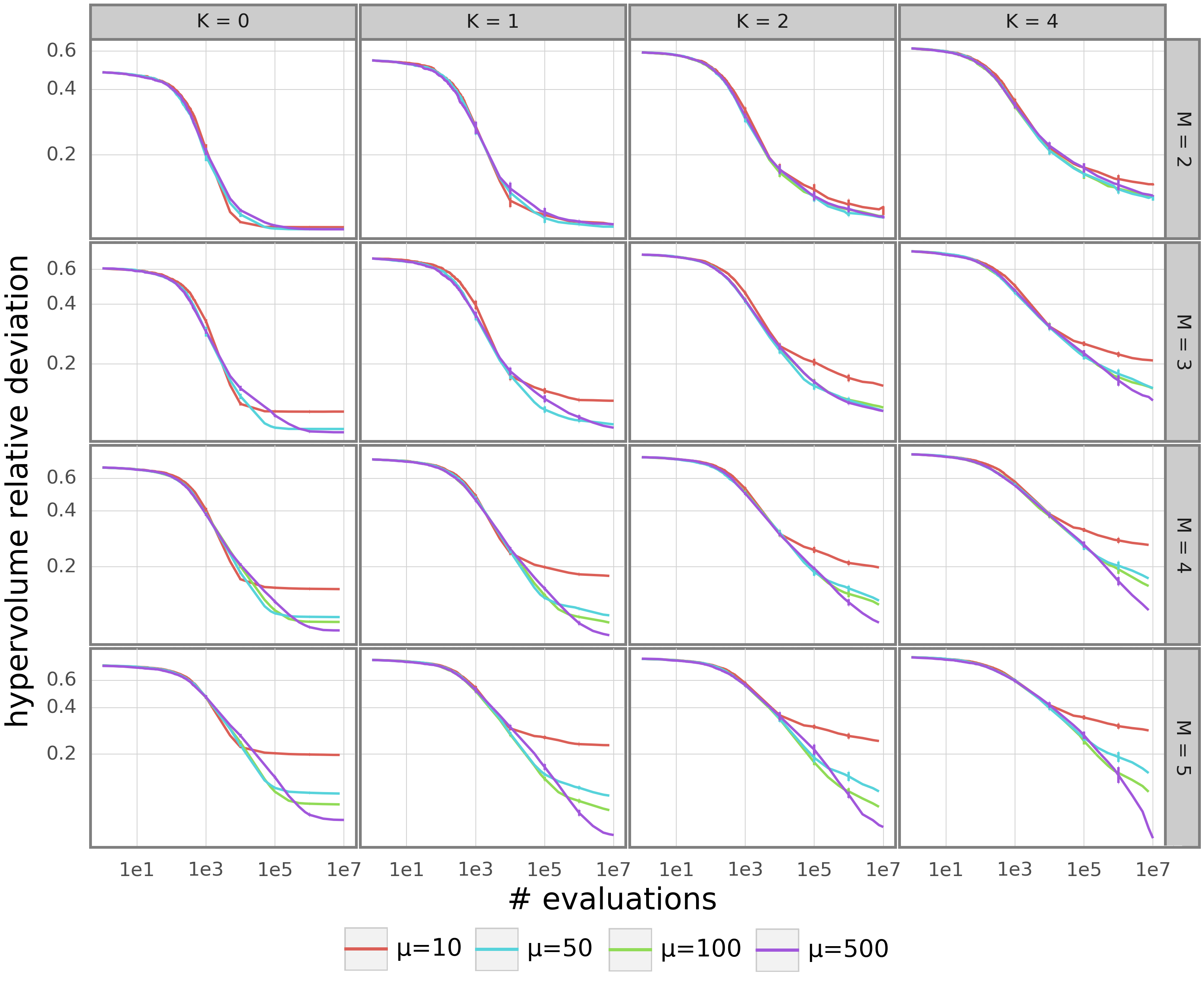}
	\vspace{-4ex}
	\caption[]{Convergence profile of the
	conventional \moead w.r.t. population size ($\mu$).%
	}
	\label{fig:comparemoeadmu}
	\vspace{-1ex}
\end{figure}

We start our analysis by studying the impact of the population size for the conventional \moead, that is \moead--($\mu$, $\mu$, \textsf{sps}$_\all$) following our terminology. In Fig.~\ref{fig:comparemoeadmu}, we show the convergence profile using different $\mu$ values for the considered instances. Recall that 
that hypervolume is measured on the external archive.

For a fixed budget, a smaller population size allows the search process to focus the computational effort 
on fewer sub-problems, hence approaching the Pareto front more quickly. By contrast, using a larger population implies more diversified solutions/sub-problems, and hence a better spreading along the Pareto front. This is typically what we observe when a small and a large budget are contrasted. 
In fact, a larger population size can be outperformed by a smaller one for relatively small budgets, especially when the problem is quite smooth~($K \leqslant 1$) and the number of objectives relatively high ($M \geqslant 3$). Notice also that it is not straightforward to quantify what is meant by a `small' population, depending on the problem difficulty. For 
a linear bi-objective problem ($M=2$, $K=0$), a particularly small population size of $\mu =10$ is sufficient to provide a relatively high accuracy. However, for quadratic many-objective problems ($M \geqslant 4$, $K=1$), a small population size of $\mu = 10$ (resp. $\mu = 50$) is only effective up to a budget of about $10^4$ (resp.~$10^5$) evaluations. 

To summarize, it appears that the approximation quality depends both on the problem characteristics and on the available budget. For small budgets, a small population size is to be preferred. However, as the available budget grows, and as the problem difficulty increases in terms of ruggedness and number of objectives, a larger population performs better. 
These first observations suggest that the anytime behavior of \moead can be improved by more advanced selection strategy, allowing to avoid wasting resources in processing a large number of sub-problems at each iteration, as implied by the conventional \textsf{sps}$_\all$ strategy which iterates over \emph{all} sub-problems.
This is further analyzed next.%

\subsection{Impact of the Sub-Problem Selection Strategy}

In order to fairly compare the different selection strategies, we analyze the impact of $\lambda$, i.e., the number of selected sub-problems, independently for each strategy. It is worth-noticing that \emph{both} the value of $\lambda$ and the selection strategy impact the probability of selecting a weigh vector. Our results are depicted in Fig.~\ref{fig:testalphafacets} for \spsdra and \spsrnd, for different budgets and on a representative subset of instances.
Other instances are not reported due to space restrictions. 
The main observation is that the best setting for $\lambda$ depends on the considered budget, on the instance type, and on the sub-problem selection strategy itself.%

\subsubsection{Impact of $\lambda$ on \spsrnd.} 
\begin{figure}[!t]
	\centering
	\includegraphics[width=1\linewidth]{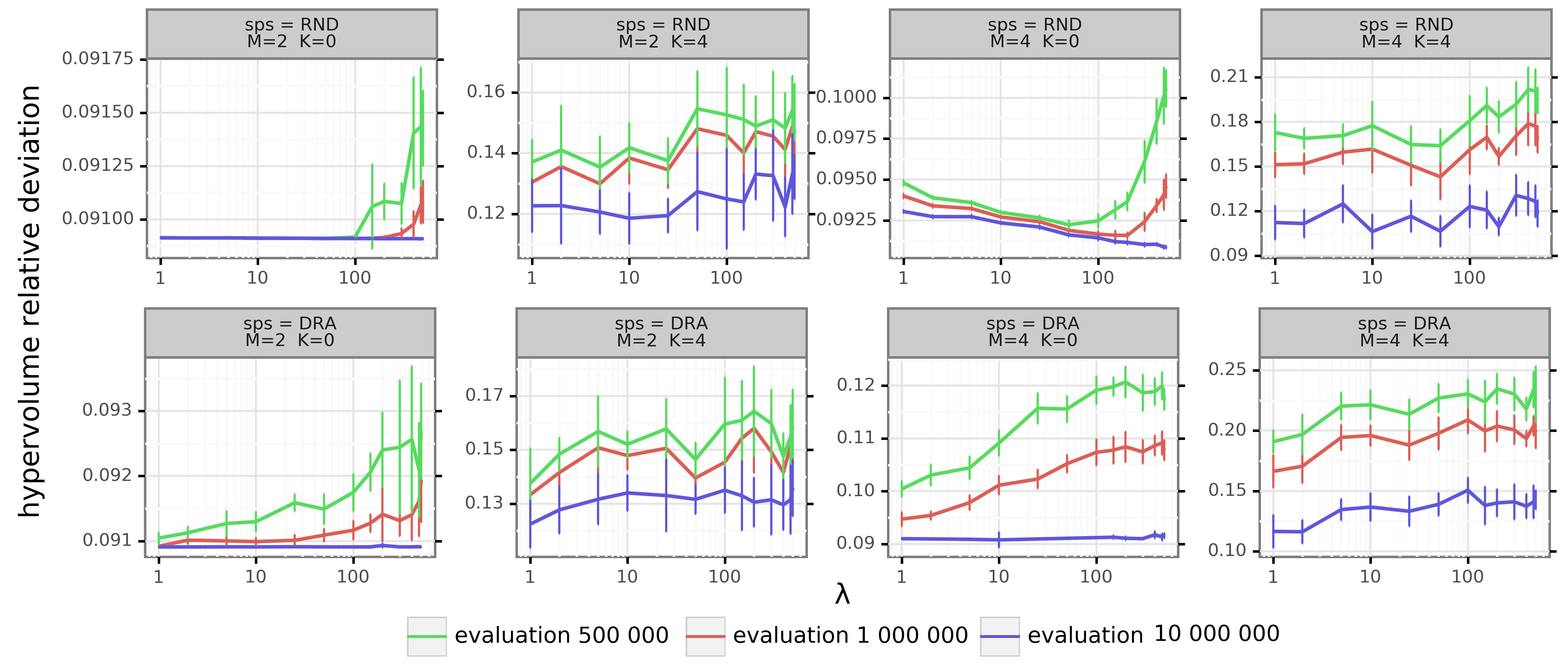}
	\vspace{-4ex}
	\caption{
	Quality vs.
	number of selected sub-problems ($\lambda$) 
	w.r.t. budget ($\mu=500$).}
	\label{fig:testalphafacets}
	\vspace{-2ex}
\end{figure}

For the random strategy \spsrnd (Fig.~\ref{fig:testalphafacets}, top), and for smooth problems ($K=0$), a small $\lambda$ value is found to perform better for a small budget. As the available budget grows, the $\lambda$ value providing the best performance starts to increase until it reaches the population size~$\mu$. 
In other words, for small budgets one should select very few sub-problems at each generation, whereas for large budgets selecting all sub-problems at each generation, as done in the standard \moead, appears to be a more reasonable choice.
However, this tendency only holds for smooth many-objective problems. When the ruggedness increases, that is when the degree of non-linearity~$K$ grows, the effect of $\lambda$ changes. For the highest value of $K=4$, the smallest value of $\lambda=1$ still appears to be effective, independently of the available budget. However, the difference with a large $\lambda$ value is seemingly less pronounced, especially for a relatively large budget, and the effect of $\lambda$ seems to decay as the ruggedness increases. Notice also that for the `easiest' problem instance (with $K=0$ and $M=2$), it is only for a small budget or for a high $\lambda$ value that we observe a loss in performance. We attribute this to the fact that, when the problem is harder, search improvements are scarce within all sub-problems, it thus makes no difference to select few or many of them at each generation. By contrast, when the problem is easier, it is enough to select fewer sub-problems, as a small number of improving offspring solutions are likely sufficient to update the population.%

\subsubsection{Impact of $\lambda$ on \spsdra.} The impact of $\lambda$ appears to be different when analyzing the \spsdra strategy (Fig.~\ref{fig:testalphafacets}, bottom). In fact, the effect of $\lambda$ seems relatively uniform, and its optimal setting less sensitive to the available budget and instance type. More precisely, the smallest value of $\lambda=1$ is always found to perform better, while an increasing $\lambda$ value leads to a decrease in the overall approximation quality. We attribute this to the adaptive nature of \spsdra, for which the probability of selecting non-interesting sub-problems is smaller for lower $\lambda$ values.
Interestingly, in the original setting of \moeaddra~\cite{subprobEquallyImportant2016}, from which \spsdra is extracted, the number of selected sub-problems is fixed to $\mu/5$. Not only we found that this setting can be sub-optimal, but it can actually be substantially outperformed by a simple setting of $\lambda=1$.%





\subsubsection{\spsall \textit{vs.} \spsdra \textit{vs.} \spsrnd} 
\begin{figure}[!t]
	\centering
	\includegraphics[width=1\linewidth]{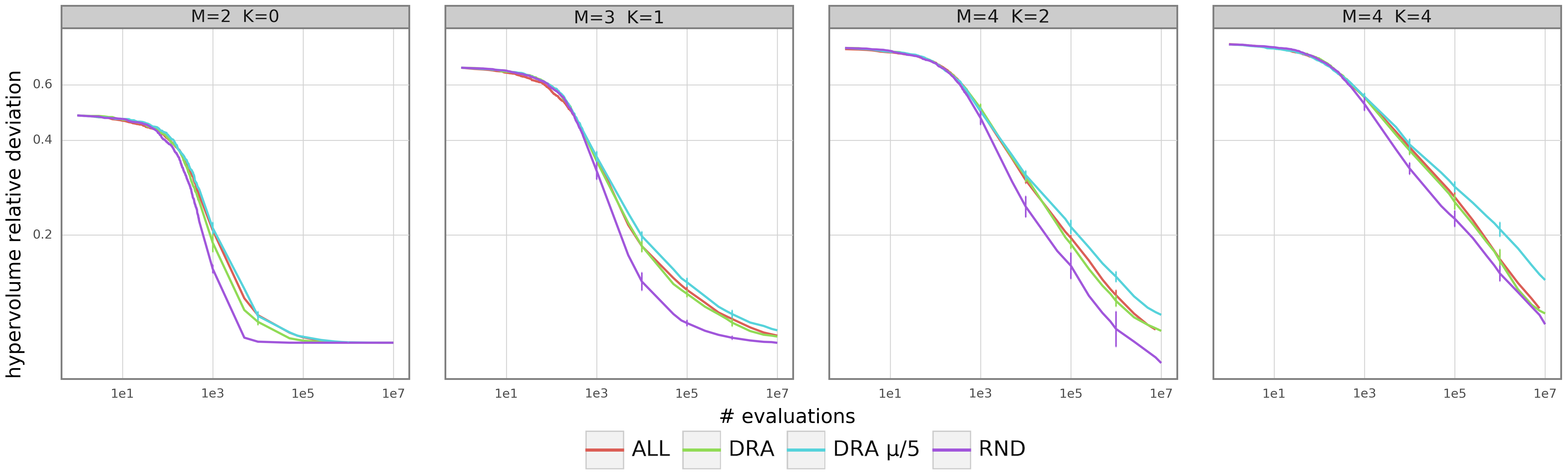}
	\vspace{-4ex}
	\caption{Convergence profile 
	of MOEA/D--($\mu$, $\lambda$, \sps)
	w.r.t. sub-problem selection strategy 
	($\mu = 500$; $\lambda = 500$ for \spsall, 
	$\lambda \in \set{1,\mu/5}$ for \spsdra, and
	$\lambda = 1$ for \spsrnd).
	}
	\label{fig:compare_sps}
	\vspace{-2ex}
	\end{figure}

Having gained insights about the effect of $\lambda$ for the different selection strategies, we can fairly analyze their relative performance by using their respective optimal setting for $\lambda$. We actually show results with $\lambda=1$ for both \spsdra and \spsrnd. Although this setting was shown to be optimal for \spsdra, it only provides a reasonably good (but sub-optimal) performance in the case of the simple random \spsrnd strategy, for which other $\lambda$ values can be even more efficient. Our results are shown in Fig.~\ref{fig:compare_sps} for a subset of instances. The \spsall strategy, corresponding to the conventional \moead~\cite{ZL07}, and \spsdra with \mbox{$\lambda=\mu/5$}, corresponding to \moeaddra~\cite{subprobEquallyImportant2016}, are also included. We can see that the simple random selection strategy \spsrnd has a substantially better anytime behavior. 
In other words, selecting a single sub-problem at random is likely to enable identifying a high-quality approximation set more quickly, for a wide range of budgets, and independently of the instance type. 

Pushing our analysis further, the only situation where a simple random strategy is outperformed by the conventional \moead or by a \moeaddra setting using an optimal $\lambda$ value is essentially for the very highest budget ($10^7$~evaluations) and when the problem is particularly smooth ($K=0$). This can be more clearly observed in Table~\ref{tab:compare_sps}, where the relative approximation quality of the different strategies are statistically compared for different budgets. Remember however that these results are for $\lambda=1$, which is shown to be an optimal setting for \spsdra, but not necessarily for \spsrnd where higher $\lambda$ values perform better.

\newcolumntype{C}[1]{>{\centering\arraybackslash}p{#1}}

\begin{table}[!t]
	\centering
\caption{Ranks and average \hvrd value (between brackets, in percentage) obtained by the different \sps strategies after $\set{10^4, 10^5, 10^6, 10^7}$ evaluations (a lower value is better). Results for \spsrnd and \spsdra are for $\lambda=1$. For each budget and instance, a rank of $c$ indicates that the corresponding strategy was found to be significantly outperformed by $c$ other strategies w.r.t. a Wilcoxon statistical test at a significance level of $0.05$.
Ranks in bold correspond to approaches that are not significantly outperformed by any other, 
and
the underlined \hvrd value corresponds to the best approach in average.\vspace{-5ex}}
	\label{tab:compare_sps}
$$
\hspace{-0.5ex}\begin{array}{@{}C{0.32cm}C{0.32cm}|ccc|ccc|ccc|ccc@{}}
		\toprule
		& &
		\multicolumn{3}{c|}{10^4 \textrm{evaluations}} & \multicolumn{3}{c|}{10^5 \textrm{evaluations}} & \multicolumn{3}{c|}{10^6 \textrm{evaluations}} &
		\multicolumn{3}{c}{10^7 \textrm{evaluations}}                               \\ \cmidrule(l){3-14} 
		\textrm{M} & \textrm{K} & \textrm{\spsall} & \textrm{\spsdra}     & \textrm{\spsrnd}                 & \textrm{\spsall} & \textrm{\spsdra}           & \textrm{\spsrnd}                 & \textrm{\spsall}              & \textrm{\spsdra}           & \textrm{\spsrnd}                 & \textrm{\spsall}              & \textrm{\spsdra}                    & \textrm{\spsrnd}                 \\ \midrule
		\multirow{4}{*}{2}       
		& 0      & 2_{(11.2)}     & 1_{(10.5)}   &  \mathbf{0}_{({\underline{09.1}})}   & 2_{(09.4)}   & 1_{(09.3)}    & \mathbf{0}_{({\underline{09.0}})} 		     & 2_{(09.1)}     	 & \mathbf{0}_{(09.0)}        & \mathbf{0}_{({ \underline{09.0}})}             			& \mathbf{0}_{(09.0)}     & \mathbf{0}_{({\underline{09.0}})} 				& 2_{(09.0)}                \\ 
		& 1      & 1_{(14.0})  & 1_{(13.2)} & \mathbf{0}_{({\underline{11.4}})}          & 1_{(10.9)}   & 1_{(10.5)}     & \mathbf{0}_{({\underline{09.5}})} 	         & 2_{(09.8)}    	 & 1_{(09.6)}        & \mathbf{0}_{({\underline{09.2}})}            				& 2_{(09.5)}                & 1_{(09.4)}           & \mathbf{0}_{({\underline{09.2}})}          \\ 
		& 2      & 1_{(17.2)}   & \mathbf{0}_{(15.6)} &  \mathbf{0}_{(\underline{{15.4}})}         & 1_{(13.0)}   & \mathbf{0}_{(12.5)}        & \mathbf{0}_{({\underline{ 11.7}})}       & \mathbf{0}_{(11.3)}     	& \mathbf{0}_{(10.9)}        	& \mathbf{0}_{({ \underline{10.6}})}         			 & \mathbf{0}_{(10.2)}                & \mathbf{0}_{(10.1)}          & \mathbf{0}_{({ \underline{09.8}})}          \\ 
		& 4      & 2_{(22.1)}   & \mathbf{0}_{(19.5)} & \mathbf{0}_{({ \underline{18.9}})}         & 1_{(17.5)}   & \mathbf{0}_{({ \underline{14.9}})}    & \mathbf{0}_{(16.0)}            & 2_{(14.6)}    	 & \mathbf{0}_{(13.3)}        	& \mathbf{0}_{({ \underline{13.0}})}           		   & \mathbf{0}_{(13.1)}                & \mathbf{0}_{({ \underline{12.0}})}          & \mathbf{0}_{(12.2)}                \\ \midrule
		\multirow{4}{*}{3}      
		& 0      & 1_{(15.1)}   & 1_{(15.0)} & \mathbf{0}_{({\underline{10.2}})}           & 1_{(11.0)}   & 1_{(10.9)}    & \mathbf{0}_{({\underline{09.2}})}		    	& \mathbf{0}_{(09.1)}      & \mathbf{0}_{(09.1)}        	& \mathbf{0}_{({ \underline{09.1}})}            	 		  & \mathbf{0}_{({\underline{09.0}})} 			& \mathbf{0}_{(09.0)}                & 2_{(09.1)}                 \\ 
		& 1     & 1_{(18.4)}   & 1_{(18.4)} & \mathbf{0}_{({\underline{14.2}})} 			& 1_{(13.3)}   & 1_{(13.0)}     & \mathbf{0}_{({\underline{10.5}})} 		   		& 1_{(10.8)}        & 1_{(10.5)}        & \mathbf{0}_{({\underline{09.4}})}            			 & 1_{(09.5)}                & 1_{(09.5)}                & \mathbf{0}_{({\underline{09.0}})} \\ 
		& 2      & 1_{(24.4)}   & 1_{(23.4)} & \mathbf{0}_{({\underline{20.6}})} 			& 1_{(16.3)}   & \mathbf{0}_{(16.1)}    & \mathbf{0}_{({\underline{14.7}})}			 & 2_{(12.8)}         & 1_{(12.3)}        & \mathbf{0}_{({\underline{11.1}})}           		     & 1_{(11.7)}                & 1_{(11.1)}                & \mathbf{0}_{({\underline{09.4}})} \\ 
			& 4      & 1_{(31.0)}   & \mathbf{0}_{(29.9)} & \mathbf{0}_{({ \underline{28.0}})}          & \mathbf{0}_{(22.7)}   & \mathbf{0}_{({ \underline{20.5}})} & \mathbf{0}_{(21.4)}             & \mathbf{0}_{(16.7)}          & \mathbf{0}_{({ \underline{15.4}})}       & \mathbf{0}_{(15.6)}            	  	 & \mathbf{0}_{(13.6)}                & \mathbf{0}_{(13.0)}                & \mathbf{0}_{({ \underline{12.2}})}          \\ \midrule
		\multirow{4}{*}{4}       
		& 0      & 1_{(20.5)}   & 1_{(20.5)} & \mathbf{0}_{({\underline{13.2}})} 		& 1_{(12.9)}   & 1_{(12.8)}        & \mathbf{0}_{({\underline{09.9}})} 			& \mathbf{0}_{(09.4)}            & \mathbf{0}_{(09.4)}        & \mathbf{0}_{({ \underline{09.3}})}       	 		  & \mathbf{0}_{({ \underline{09.0}})}          & \mathbf{0}_{(09.0)}                & 2_{(09.2)}               \\ 
		& 1      & 1_{(25.0)}   & 1_{(24.9)} & \mathbf{0}_{({\underline{18.5}})} 		& 1_{(15.2)}   & 1_{(15.4)}        & \mathbf{0}_{({\underline{11.8}})} 			& 1_{(09.9)}            & 1_{(09.8)}        & \mathbf{0}_{({\underline{08.8}})}       			  & 1_{(08.4)}                 & 1_{(08.5)}                & \mathbf{0}_{({\underline{07.9}})} \\ 
		& 2      & 1_{(29.8)}   & 1_{(30.6)} & \mathbf{0}_{({\underline{24.7}})} 		& 1_{(19.5)}   & 1_{(18.7)}        & \mathbf{0}_{({\underline{16.0}})} 			& 1_{(12.9)}            & 1_{(12.3)}        & \mathbf{0}_{({\underline{10.1}})}       	 		  & 1_{(09.8)}                 & 1_{(09.9)}                & \mathbf{0}_{({\underline{07.9}})} \\ 
		& 4      & 1_{(38.0)}   & 1_{(37.1)} & \mathbf{0}_{({\underline{32.5}})} 		& 1_{(26.4)}   & 1_{(25.4)}        & \mathbf{0}_{({\underline{22.5}})} 			& 1_{(16.8)}            & 1_{(16.6)}        & \mathbf{0}_{({\underline{15.1}})}          		  & \mathbf{0}_{(11.6)}                 & \mathbf{0}_{(11.3)}                &\mathbf{0}_{( { \underline{10.5}})}          \\ \midrule
		\multirow{4}{*}{5}      
		& 0      & 2_{(26.3)}   & 1_{(25.2)} & \mathbf{0}_{({\underline{15.4}})} 		& 1_{(14.1)}   & 1_{(14.7)}        & \mathbf{0}_{({\underline{10.2}})} 			& \mathbf{0}_{({\underline{08.0}})} 			& 1_{(08.3)}        	& 2_{(08.5)}             	  & \mathbf{0}_{({ \underline{07.4}})}          & \mathbf{0}_{(07.5)}                & 2_{(08.0)}                \\ 
		& 1      & 1_{(29.9)}   & 1_{(29.9)} & \mathbf{0}_{({\underline{21.5}})} 		& 1_{(16.5)}   & 1_{(17.1)}        & \mathbf{0}_{({\underline{13.0}})} 			& 1_{(08.3)}               & 1_{(08.5)}        & \mathbf{0}_{({\underline{07.7}})}            	  & \mathbf{0}_{({ \underline{05.9}})}          & \mathbf{0}_{(06.1)}                & 1_{(06.1)}                \\ 
		& 2      & 1_{(35.2)}   & 1_{(34.1)} & \mathbf{0}_{({\underline{28.1}})} 		& 1_{(21.4)}   & \mathbf{0}_{(20.0)}        &  \mathbf{0}_{(\underline{17.6})}           & \mathbf{0}_{(10.9)}            & \mathbf{0}_{(10.5)}        	& \mathbf{0}_{({ \underline{09.7}})}         		 & \mathbf{0}_{(06.8)}              & \mathbf{0}_{(06.2)}                & \mathbf{0}_{({\underline{05.3}})} \\ 
		& 4      & 1_{(41.6)}   & 1_{(40.7)} & \mathbf{0}_{({\underline{35.5}})} 		& 1_{(26.5)}   & 1_{(26.9)}        & \mathbf{0}_{({\underline{24.1}})} 			& \mathbf{0}_{(14.7)}              & \mathbf{0}_{(15.1)}        & \mathbf{0}_{( { \underline{14.3}} )}        		 & \mathbf{0}_{({ \underline{06.0}})}        	  & \mathbf{0}_{(07.4)}                & \mathbf{0}_{(07.4)}                 \\ \bottomrule
	\end{array}
	$$\vspace{-4ex}
\end{table}

\subsection{Robustness of MOEA/D--($\mu$, $\lambda$, \spsrnd) w.r.t. $\mu$ and~$\lambda$}

\begin{figure}[!t]
	\centering
	\includegraphics[width=1\linewidth]{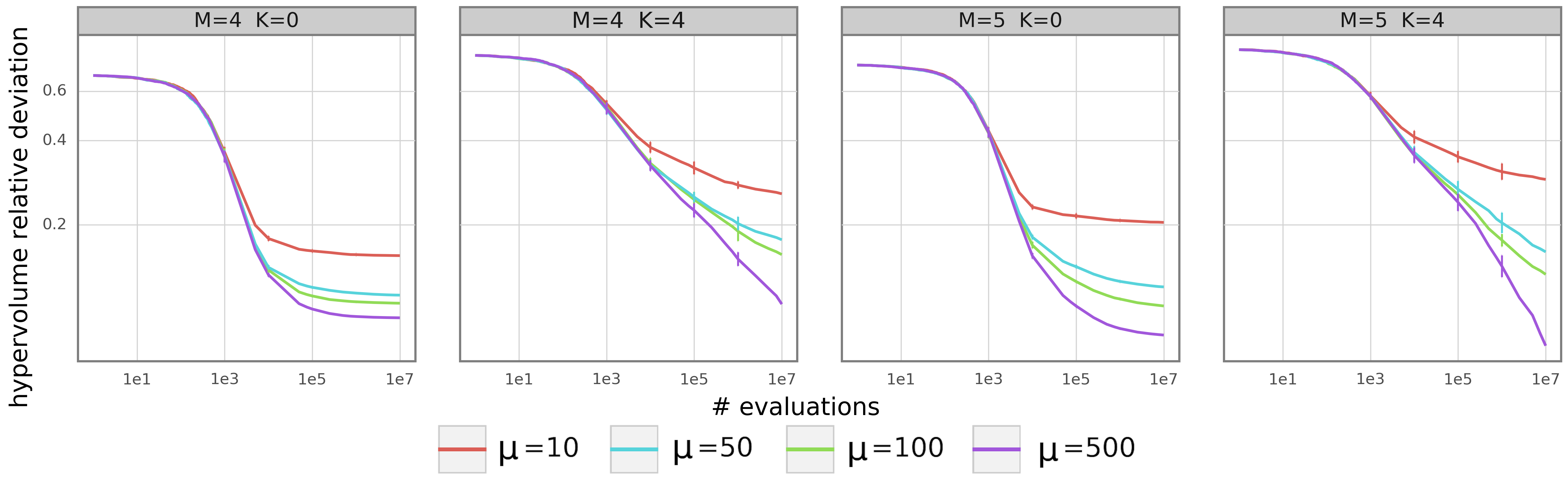}
	\vspace{-4ex}
	\caption{
	Convergence profile of 
	MOEA/D--($\mu$, $1$, \spsrnd) w.r.t. population size ($\mu$).%
	}
	\label{fig:comparemuwithalpha1}
	\vspace{-2ex}
\end{figure}

In the previous section, the population size was fixed to the highest value of $\mu=500$. However, we have shown in Section~\ref{sec:popmoead} that the anytime behavior of the conventional \moead can be relatively sensitive to the setting of $\mu$, in particular for some instance types. Hence, we complement our analysis by studying the sensitivity of the \spsrnd strategy, which was found to have the best anytime behavior overall, w.r.t the population size $\mu$. Results for \spsrnd with $\lambda=1$ are reported in Fig.~\ref{fig:comparemuwithalpha1}. In contrast with the \spsall strategy from the conventional \moead reported in Fig.~\ref{fig:comparemoeadmu}, we can clearly see that the anytime behavior underlying \spsrnd is much more stable. In fact, the hypervolume increases with~$\mu$, independently of the considered budget and instance type. Notice also that when using small $\mu$ values, convergence occurs much faster for smooth problems ($K=0$) compared against rugged ones ($K=4$). This means that a larger population size $\mu$, combined with a small value of $\lambda$, shall be preferred.

From a more general perspective, this observation is quite insightful since it indicates that, by increasing the number of weight vectors, one can allow for a high-level structure of the population, being eventually very large. Notice also that such a data structure can be maintained very efficiently in terms of CPU time complexity, given the scalar nature of \moead. This is to contrast with, e.g., dominance-based EMO algorithms, where maintaining a large population may be computationally intensive, particularly for many-objective problems. Having such an efficient structure, the issue turns out to select some sub-problems from which the (large) population is updated. A random strategy for sub-problem selection is found to work arguably well. However, in order to reach an optimal performance, setting up the number of sub-problems~$\lambda$ might require further configuration issues. Overall, our analysis, reveals that a small~$\lambda$ value, typically ranging from $1$ to $10$, is recommended for relatively rugged problems, whereas a large value of $\lambda$ should be preferred for smoother problems.

\section{Conclusions and Perspectives}
\label{sec:conclusion}


In this paper, we reviewed the design principles of the 
\moead framework by providing a high-level, but more precise, reformulation 
$(\mu + \lambda)$ scheme from evolutionary computation. 
We analyzed the role of three design components: the population size~($\mu$), the number of sub-problems selected 
and the strategy used for sub-problems selection~(\sps). Besides systematically informing about the combined effect of these components on the performance profile of the search process as a function of problem difficulty in terms of ruggedness and objective space dimension, our analysis opens new challenging questions on the design and practice of decomposition-based EMO algorithms. 

Although we are now able to derive a parameter setting recommendation according to the general properties of the problem at hand, such properties might not always be known beforehand by the practitioner, and other properties might be considered as well. For instance, one obvious perspective would be to extend our analysis to the continuous domain. More importantly, an interesting research line would be to infer the induced landscape properties in order to learn the `best' parameter setting, either off-line or on-line. 
This would not only avoid the need of additional algorithm configuration 
efforts, but it could also lead to an even better anytime behavior. One might for instance consider an adaptive setting where the values of $\mu$, $\lambda$, and \sps are adjusted 
according to the search behavior 
observed over different generations. Similarly, we believe that considering a problem where the objectives expose some degree of heterogeneity, e.g., in terms of solving difficulty, is worth investigating. 
In that case, the design of an accurate \sps strategy is certainly a key issue.
More generally, we advocate for a more systematic analysis of such considerations for improving our fundamental understanding of the design issues behind \moead and EMO algorithms in general, of the key differences between EMO algorithm classes, and of their success in solving challenging multi- and many-objective optimization problems.

\subsubsection{Acknowledgments}
This work was supported by the French national research agency (ANR-16-CE23-0013-01) and the Research Grants Council of Hong Kong
(RGC Project No. A-CityU101/16).

%
%
%
\bibliographystyle{splncs04}
\bibliography{biblio}

%

\end{document}